\documentclass[10pt,letterpaper]{article}

\usepackage{cogsci}
\usepackage{amsmath}
\usepackage{xurl}

\cogscifinalcopy 

\usepackage{pslatex}
\usepackage{apacite}
\usepackage{graphicx}
\usepackage{float} 

\title{Modeling Understanding of Story-Based Analogies Using Large Language Models}
\author{{\large \bf Kalit Inani*, Keshav Kabra*, Vijay Marupudi, Sashank Varma }\\
\{kinani3, keshav.kabra, vijaymarupudi, varma\}@gatech.edu \\
  Georgia Institute of Technology}
\begin{document}
\maketitle
\begin{abstract}
Recent advancements in Large Language Models (LLMs) have brought them closer to matching human cognition across a variety of tasks. How well do these models align with human performance in detecting and mapping analogies? Prior research has shown that LLMs can extract similarities from analogy problems but lack robust human-like reasoning. Building on \citeA{Webb2023}, the current study focused on a story-based analogical mapping task and conducted a fine-grained evaluation of LLM reasoning abilities compared to human performance. First, it explored the semantic representation of analogies in LLMs, using sentence embeddings to assess whether they capture the similarity between the source and target texts of an analogy, and the dissimilarity between the source and distractor texts. Second, it investigated the effectiveness of explicitly prompting LLMs to explain analogies. Throughout, we examine whether LLMs exhibit similar performance profiles to those observed in humans by evaluating their reasoning at the level of individual analogies, and not just at the level of overall accuracy (as prior studies have done).
Our experiments include evaluating the impact of model size (8B vs. 70B parameters) and performance variation across state-of-the-art model architectures such as GPT-4 and LLaMA3. 
This work advances our understanding of the analogical reasoning abilities of LLMs and their potential as models of human reasoning.

\textbf{Keywords:} 
analogical reasoning; story-based analogies; Large Language Models; LLMs; prompt engineering
\end{abstract}

\section{Introduction}
Analogical reasoning is a fundamental aspect of human cognition. When faced with novel problems, humans can come up with creative solutions by drawing on similarities to and differences from familiar problems. Recent advancements in Large Language Models (LLMs) have expanded their ability to reason, infer, and deduce. Prior research has investigated their performance in zero-shot reasoning: solving problems without prior exposure during training. 

In this study, we investigate the performance of LLMs on a classic set of story analogy problems \cite{Gentner1993}. We also consider the models' robustness to answer ordering \cite{Lewis2024EvaluatingTR}. Finally, we evaluate the possible improvement in model-human alignment with an enhanced 2-step prompting approach, to better simulate the approach humans might take. Throughout, we explore multiple different LLM architectures and varying numbers of LLM training parameters. 

By building on and comparing our findings with those of previous studies, notably \citeA{Webb2023} and \citeA{Lewis2024EvaluatingTR}, we contribute to the expanding understanding of how well LLMs match human cognition in terms of analogical reasoning. This comprehensive analysis provides insights into the reasoning abilities of current state-of-the-art AI/ML/NLP models.

\subsection{Related Work}

\citeA{Webb2023} compared the performance of LLMs, notably OpenAI's GPT-3 \cite{Brown2020LanguageMA}, with human cognition across a variety of analogical reasoning tasks: story analogies, four-term verbal analogies, letter string analogies, and digit matrices. GPT-3 demonstrated remarkable zero-shot analogical reasoning abilities across these four tasks, frequently matching or outperforming the overall task accuracy of humans. Nonetheless, some limitations were observed. Notably, for the story analogy task that is the focus of the current study, GPT-3's performance (0.67) was noticeably poorer than that of humans (0.85). Both GPT-3 and humans demonstrated sensitivity to higher-order relations on this task. However, GPT-3 had more difficulty with far analogies (cross-domain comparisons) than with close analogies (within-domain comparisons). Thus, while GPT-3 exhibits emergent analogical reasoning skills, it still performs below the level of humans on items that call for intricate causal reasoning and cross-domain comparisons. Similarly, \citeA{Sourati-etal-2024-arn} performed a systematic evaluation of narrative-level system mappings. They show that even though LLMs are able to recognize near analogies, they struggle with far analogies in a zero-shot setting. These results motivate the need for further investigation into the ways in which LLMs comprehend narratives and make sense of intricate causal chains, especially in cross-domain settings.

\citeA{Lewis2024EvaluatingTR} found that GPT models face challenges in zero-shot analogy formation, especially for story analogies. In particular, these researchers found that these models are vulnerable to answer-order effects: GPT-4's accuracy was 89\% when the correct answer was shown first, but it decreased to 61\% when it was shown second. In contrast, human performance is unaffected by the order of answers (i.e., when the correct one is presented). Their study also showed differential effects of paraphrasing in models versus humans. In general, performance decreases when the correct target story was paraphrased to reduce surface similarities with the source story. GPT-4's accuracy dropped from 86\% on original stories to 72\% on paraphrased stories. Human accuracy also dropped, but half as much, from 78\% to 70\%. This suggests that GPT-4 is more susceptible to paraphrasing effects than humans. These results show the importance of assessing the performance of ML models not only for accuracy but also for robustness. They imply that although the analogical reasoning of GPT models has advanced significantly, it still falls short of human cognition in terms of flexibility and generalizability, especially in zero-shot situations.

\subsection{Research Questions}

Despite the important studies of \citeA{Webb2023} and \citeA{Lewis2024EvaluatingTR}, a number of questions remain about the analogical reasoning abilities of LLMs and their alignment with human cognition. The current study focuses on a signature of human analogical reasoning, its sensitivity to the higher-order causal relations between events. These relations can serve as the basis of analogies \cite{Gentner1993}. Here, we ask if LLMs are also sensitive to causal structure for story-based analogies. In particular, we address the following research questions:
\begin{enumerate}
  \item  What is the semantic representation of stories in encoder-based LLMs? These models are often used to extract vector representations of text. Specifically, to what extent might the vector representations of source stories show greater similarity to analogically related target (correct) stories compared to unrelated distractor (incorrect) stories?
  \item Can the analogical reasoning performance of LLMs and their alignment to human performance be improved through a prompting strategy that uses self-generated hints?
  \item How close is the alignment between models and humans, not just at coarse level of overall accuracy but at a finer-grained level of individual items? Do they find the same analogies difficult?
  \item How do the results vary as a function of model size and architecture? 
\end{enumerate}



\section{Method}

\subsection{Materials}

We used the story analogy problems from \citeA{Gentner1993}. The problems were retrieved from the repository mentioned in \citeA{Webb2023}. The stimuli consist of 18 problems, where each problem constitutes a source story and two potential target stories. Each target story shares first-order relations with the source but involves different entities and higher-order relations. However, only the correct target story shares the same causal structure as the source (referred to as ‘true analogy’). Here is an example item from the stimulus set: \\

Source story: \textit{Once there was a teacher named Mrs. Jackson who wanted a salary increase. One day, the principal said that he was increasing his own salary by 20 percent. However, he said, there was not enough money to give the teachers a salary increase. When Mrs. Jackson heard this she became so angry that she decided to take revenge. The next day, Mrs. Jackson used gasoline to set fire to the principal's office. Then she went to a bar and got drunk.\\} 

True Analogy: \textit{McGhee was a sailor who wanted a few days of vacation on land. One day, the captain announced that he would be taking a vacation in the mountains. However, he said, everyone else would have to remain on the ship.
After McGhee heard this he became so upset that he decided to get revenge. Within an hour McGhee blew up the captain's cabin with dynamite.\\}

False Analogy: \textit{McGhee was a sailor who wanted a few days of vacation on land. One day McGhee became so impatient that he tried to blow up the captain's cabin using dynamite.
After this incident, the captain announced that he would be taking a vacation in the mountains. However, he said, everyone else would have to remain on board to repair the ship.\\} 

In the above example, the source story (Mrs. Jackson --- principal) and the target analogies (McGhee --- captain) differ in the entities involved. In contrast, the target analogies share both first-order relations and entities, making them superficially similar. However, only the true target analogy shares the same higher-order relation with the source story: the idea of taking revenge.

In our experiments, we presented the models with each of the 18 problems
and prompted them to find the target story that is most analogous to
the source story. The corresponding human data were obtained from an
experimental replication of the original \citeA{Gentner1993} study by
\citeA{Webb2023}. We thank Taylor Webb for making these resources
publicly available and answering our questions about them.

\subsection{Models}

To address the first research question, we selected a model from the \textbf{BERT} family, specifically \texttt{bert-base-uncased} \cite{Devlin2019BERTPO}, to generate sentence embeddings for the source and potential target stories.  The model was pre-trained on a large corpus of English data in a self-supervised manner, with no human labeling. It was trained using two objectives, masked language modeling and next-sentence prediction. This helps the model learn essential features in sentences and generate robust embeddings. We used BERT as its bidirectional architecture provides rich contextual representations.
We used the Python library \texttt{transformers} \cite{Devlin2019BERTPO} for our implementation.

To address the other research questions, we used generative transformer models spanning two architectural families: OpenAI's \textbf{GPT-4} \cite{Achiam2023GPT4TR} and Meta's \textbf{LLaMA3.1} \cite{Dubey2024TheL3}. We used the default temperature value of 1 in these models. Both achieve high scores on an array of standard NLP benchmarks investigating language ability and reasoning. Within each architecture, we performed tests with models of varying size, i.e., numbers of parameters. 

\subsection{Tasks}

\subsubsection{Sentence embedding task}

To address the first research question -- whether the sentence embeddings of LLMs capture the causal relationships present in analogies -- we generated the sentence embeddings using BERT for the source story and for each of the target stories. Next, we evaluated the similarity of embedding for the source story with these embedding of each target story using cosine similarity as the metric. The values range between $-1$ and $1$, with larger values signaling greater semantic similarity between a pair of stories. We evaluated whether, for each analogy item, the cosine similarity between the source story and the `true analogy' story was greater than that between the source story and the `false analogy' story.





\subsubsection{Generative model task}

We perform experiments on LLMs by prompting them with the source and target stories, replicating and extending the procedure used by \citeA{Webb2023} and \citeA{Lewis2024EvaluatingTR}.

In \textbf{conventional prompting}, for each of the analogy items, the model is presented with the source story and the two target stories, denoted Story A and Story B. The model is then prompted: \\

\textit{Based on the stories, predict which of the target stories is more analogous to the source story. Just output a single-word answer: `Story A' or `Story B'.} \\

We prompt the model with the same problem across 100 different instances to average over any randomness in the results. Additionally, to avoid any ordering bias \cite{Lewis2024EvaluatingTR}, half (50) of the prompts are constructed to have `Story A' as the true analogy while the other half have `Story B' as the true analogy. We force the models to make a single answer choice (i.e., either `Story A' or `Story B') based on the prompt. For each of the 18 stories, the results are averaged over 100 runs presented to the models. 


To extend prior work and to potentially make the model more closely mimic human reasoning, we also engaged in prompt engineering. Specifically, we devised a 2-step \textbf{enhanced prompting} method that instructs the model to outline the concepts necessary to identify causality and focus on analogy-relevant aspects of the stories. In the first step, the source story and the potential target stories are presented to a generative model and the model is asked to generate hints for a model to use in the future. Here is the prompt provided to the models to generate these hints: \\

\textit{
Consider the following source story: [Text for the source story]\newline
Now, consider two target stories: \newline
Story A: [Text for the `true' target story 1] \newline
Story B: [Text for the `false' target story 2] \newline
Output a prompt with human-like self-generated hints which an LLM can use to make an inference similar to humans.
}\\

Here are the self-generated hints generated by GPT based on the above prompt for the example analogy problem presented above:\\

\textit{In the source story, Mrs. Jackson's anger toward the principal's actions motivates her to take an extreme action (setting fire to his office) as a form of revenge. Her response is clearly linked to the principal's decision not to give teachers a raise while increasing his own salary.
Similarly, McGhee in Story A is motivated by the captain's decision to take a vacation while everyone else stays on the ship. McGhee’s response to this injustice is to take an extreme action (blowing up the captain’s cabin) as a form of revenge.
In Story B, McGhee’s action of trying to blow up the captain’s cabin seems more driven by impatience or frustration, not directly in response to an unfair action, and the sequence of events feels out of order with the captain’s decision happening after the incident.}

We see that the hints, generated by the model itself without being provided the correct answer, instruct the subsequent model to focus on identifying revenge motives in the given analogy example.
5

In the second step, these hints are then appended to the original prompt and again presented to the models for inference.

\textit{Now, based on the stories and the hints, predict which of the target stories is more analogous to the source story. Just output a single-word answer: `Story A' or `Story B'.}

Similar to the previous approach, the enhanced prompt is run against the model 100 times and the accuracy score is aggregated across the $18 \times 100 = 1800$  runs.

\subsubsection{Variance across models and model sizes}

Previous research focused only on the analogical reasoning ability of OpenAI's GPT-based models \cite{Lewis2024EvaluatingTR, Webb2023}. Here, we extend the evaluation to LLaMA as well. This allows us to understand whether differences in transformer-based architectures may result in differences in reasoning performance. 

We also assess whether analogical reasoning performance improves with more parameters. For GPT, the experiments are performed over \textit{gpt-4o} \cite{gpt4} and \textit{gpt-4o-mini} \cite{gpt4o-mini}. For LLaMA, the models used are \textit{llama-3.1-8B-Instruct} (8 billion parameters) and \textit{llama-3.1-70B-Instruct} (70 billion parameters) \cite{llama3.1}. 

All models were subjected to both prompting approaches: conventional prompts and enhanced prompts.

\subsubsection{Correlation with human performance}

To evaluate how models of different architectures and varying parameters resemble human cognition, we use Pearson correlations. Specifically, for each model and each prompting approach (conventional vs. enhanced), we compute the accuracy for each of the 18 problems. We then compute the Pearson correlation between these accuracies and those of humans in the \citeA{Webb2023} replication of the \citeA{Gentner1993} experiment. This quantifies the model-human alignment.

\section{Results}

\subsection{Sentence embedding task} 

The first research question concerns the semantic representation of the source and target stories, and whether the source is more similar to the `true analogy' target than the `false analogy' target. Using BERT-based story embeddings and the cosine similarity metric, we observed an accuracy of 0.78. That is, for 78\% of the analogy items, the higher similarity was to the `true analogy' target story. This overall accuracy is less than the 84.7\% observed for humans.

\begin{figure}
\begin{center}
\includegraphics[width=0.5\textwidth]{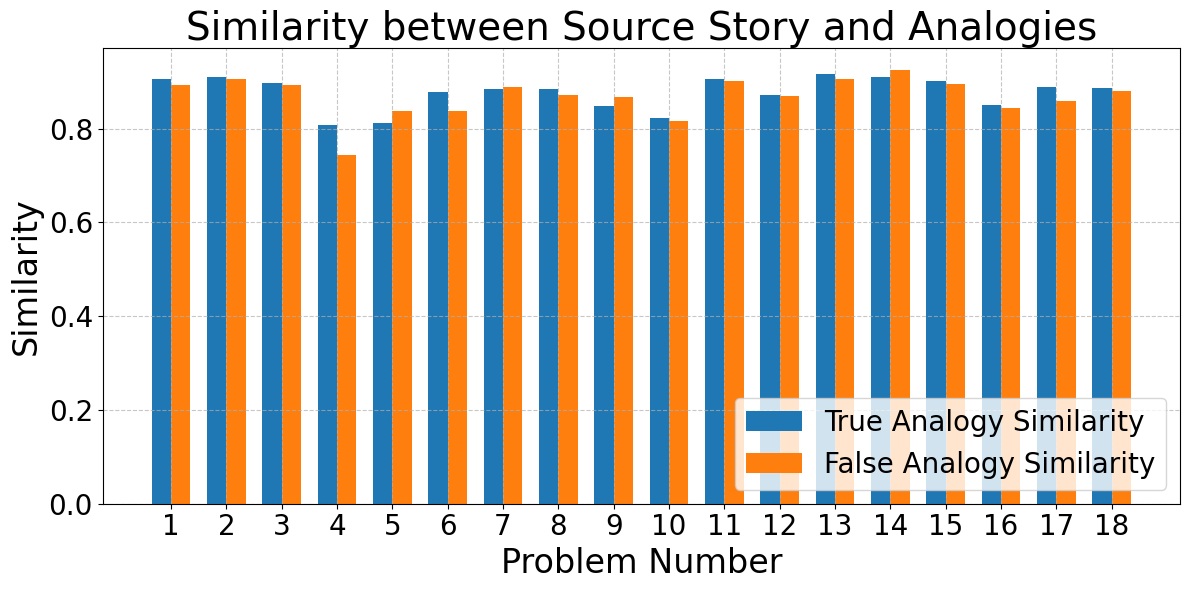}
\end{center}
\caption{Cosine similarity scores of the BERT embeddings of the source story and each of the `true analogy' and `false analogy' targets, for the 18 analogy problems.} 
\label{bert-similarity}
\end{figure}

At a finer-grained level,  Figure~\ref{bert-similarity} shows, for each analogy item, the similarity of the source story to the `true analogy' target (blue) and the `false analogy' target (red). Although the model's overall accuracy is 78\%, it is clear that the similarity scores are more comparable than distinguishable between the true and false analogies.  

\begin{figure}
\begin{center}
\includegraphics[width=0.5\textwidth]{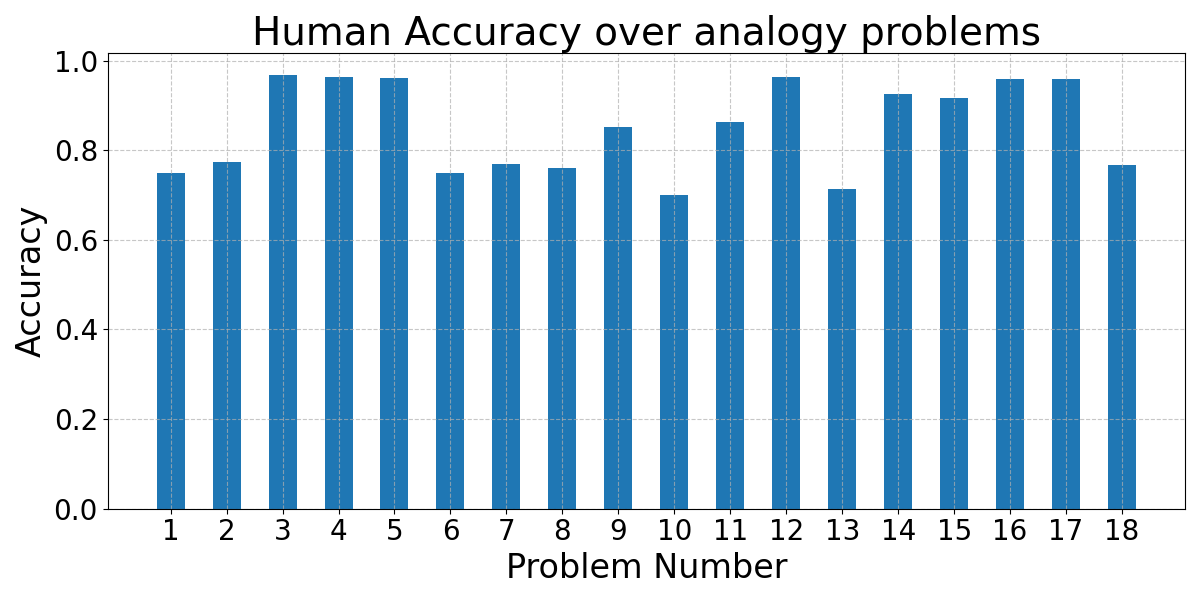}
\end{center}
\caption{Human accuracies for the 18 analogy problems.} 
\label{human-accuracy}
\end{figure}

To address the third research question, we evaluated alignment at a finer-grain level, computing the correlation at the item level between human accuracy (plotted in Figure ~\ref{human-accuracy}) and the `true analogy' similarity minus the `false analogy' similarity (plotted in Figure~\ref{bert-similarity}). The value was $r = -0.032$ ($p = 0.899$), indicating no agreement between humans and the model on whether individual analogy items were relatively easy or relatively difficult. This suggests that despite its high accuracy, the model does not follow a similar performance profile to humans.

\subsection{Generative model task}

The second research question asks whether the alignment between humans and the models can be improved through a prompting strategy that uses self-generated hints. Table \ref{gpt-llama-enhanced-prompt-performance} shows the relevant results. A number of patterns stand out.

First, with respect to the second research question, the overall performance of both the GPT-4 and Llama 3.1 models is greater for the enhanced prompt than for the conventional prompt.

Second, with respect to the fourth research question, we investigated the effect of model size. There is a clear pattern of higher model accuracy with additional model parameters: this holds for both model families and for both prompting approaches.
From Table \ref{gpt-llama-enhanced-prompt-performance}, we can see that
the larger models perform at least 15\% better than their smaller counterparts.
Llama 3.1 model with 70B parameters has a much better accuracy of 0.8538 compared to the 8B parameter model, which has an accuracy of 0.6528. Similarly, there is a nearly 15\% increase in performance from GPT-4o-mini to GPT-4o (i.e., higher parameter model). Moreover, for the larger models, the accuracies exceed those achieved by humans.

\begin{table}
\begin{center} 
\caption{Overall model accuracies as a function of model size and prompting strategy. For reference, the results for GPT-3 from \citeA{Webb2023} are also shown.} 
\label{gpt-llama-enhanced-prompt-performance} 
\vskip 0.12in
\resizebox{\columnwidth}{!}{
\begin{tabular}{ccc} 
\hline
Model    &  Conventional Prompt  & Enhanced Prompt  \\
\hline
Humans  & 0.847 & N/A \\
gpt-3 \cite{Webb2023} &   0.75 & N/A \\
\hline
gpt-4o-mini        &   0.7011 & 0.7411 \\
gpt-4o        &   0.8233 & 0.8850 \\
llama-3.1-8B-Instruct        &   0.6528 & 0.7472 \\
llama-3.1-70B-Instruct       &   0.8538 &  \textbf{0.9150} \\
\hline
\end{tabular} 
}
\end{center} 
\end{table}

\subsection{Human-Model Alignment at the Item Level}

\begin{figure}
\begin{center}
\includegraphics[width=\columnwidth]{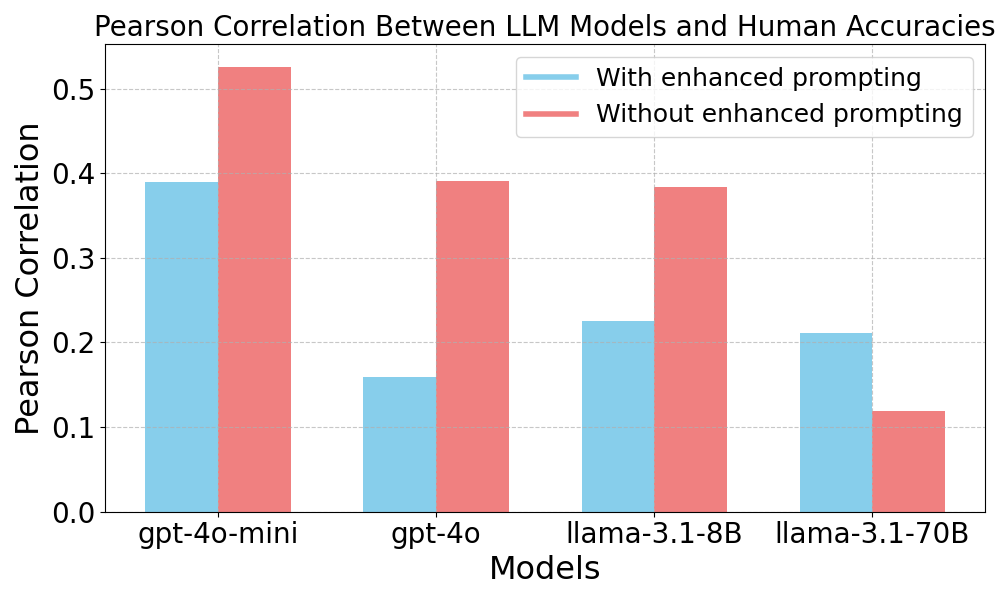}
\end{center} 
\caption{The Pearson correlation of each model's accuracy with humans over the 18 problems. Note that only GPT-4o-mini without the enhanced prompting strategy achieves a correlation significantly different than 0 ($r = 0.526, p = 0.025)$.}
\label{correlation-matrix}
\end{figure}

The third research question asks, at a finer-grain level, whether humans and the models find the same analogy items to be easy and the same ones to be difficult. We evaluated this by computing, for each model and each prompting approach the Pearson correlation between the human and model accuracies across the 18 items. The results are shown in Figure \ref{correlation-matrix}. The contrast to the overall accuracies in Table \ref{gpt-llama-enhanced-prompt-performance} is instructive. Llama 3.1-70B achieves the highest overall accuracy. However, it is the smallest model, GPT-4o-mini, that achieves the highest correlation to human performance, i.e., best tracks which items humans find easy and which ones they find difficult. Moreover, only this model combined with the conventional prompting strategy achieves a correlation significantly different than 0 ($r = 0.526, p = 0.025)$.


Another interesting difference concerns prompting. Enhanced prompting enables all models, regardless of size, to achieve higher overall accuracies than conventional prompting. However, the finding reverses for the correlations: for three of the four models/sizes, a higher correlation is achieved at the individual item level for conventional vs. enhanced prompting. The only exception to this pattern is Llama 3.1-70B, the model/size that achieves the lowest correlations overall.
The lower correlations in the enhanced prompting approach may be attributable to a ceiling effect: because the model accuracies approach the maximum possible score, their variability across items is reduced and potentially distorted. This reduces the ability to evaluate whether the models track human accuracies at the item level.

\begin{figure}[ht]
    \begin{center}
    \begin{tabular}{c} 
        \includegraphics[width=\columnwidth]{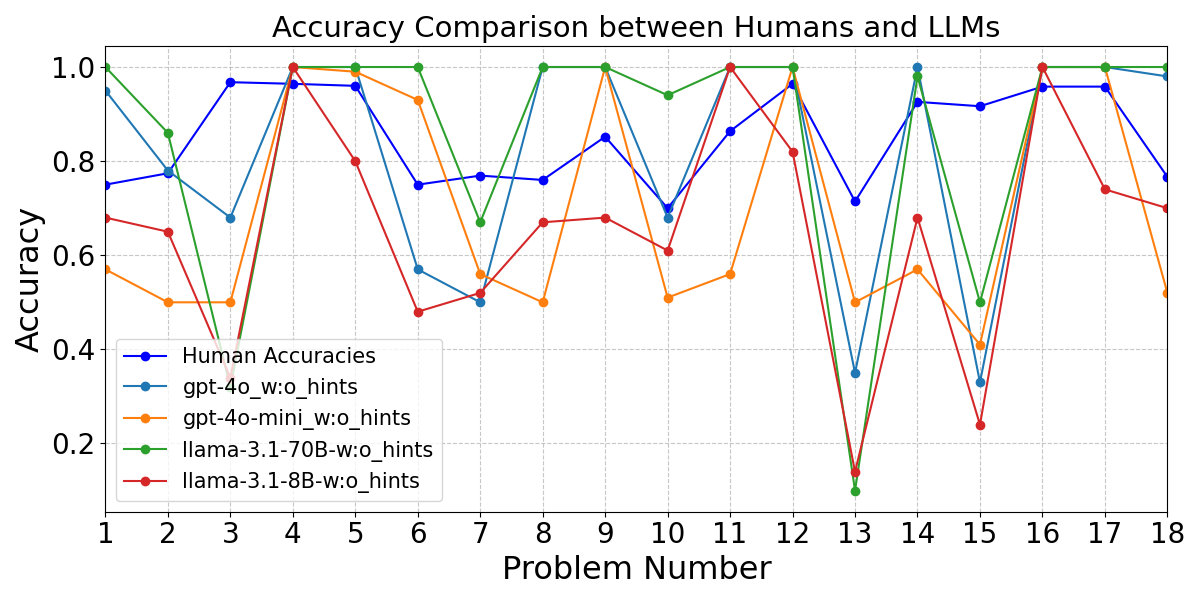} \\ 
        \\
        \includegraphics[width=\columnwidth]{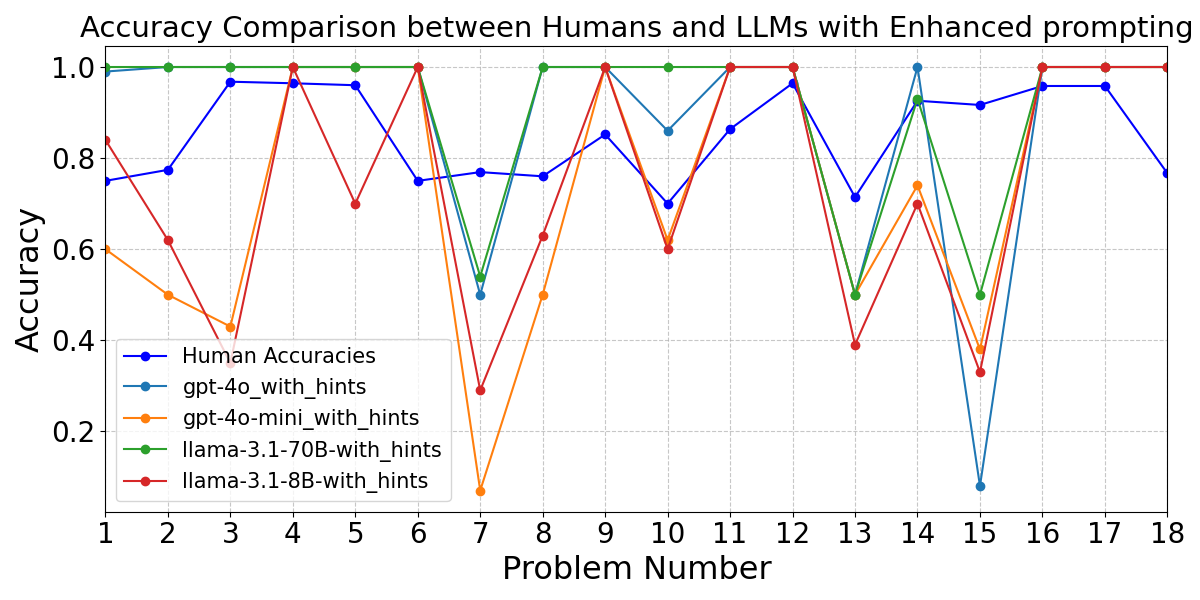} \\ 
    \end{tabular}
    \end{center}
    \caption{Accuracy comparison of human and LLM performances on each of the 18 story analogy problems, under conventional prompting (top) and enhanced prompting (bottom).}
    \label{table_model_human_improved_prompt}
\end{figure}

Finally, we took a closer look at the items where humans and the models most diverge. Figure \ref{table_model_human_improved_prompt} plots the accuracy of humans and each of the four models for each of the 18 items. Although GPT and LLaMA perform as well as or better than humans for most problems, there are a few notable exceptions: problems 3, 7, 13 and 15. To understand why the models might be having difficulty with these items, we take a closer look. Here is problem 7 of the dataset:

Source story: \textit{Percy the mockingbird spent the whole warm season chirping and twittering. When it began to get colder Percy visited a squirrel and sang a song for her, expecting to get some of the squirrel's sunflower seeds in return. However, the squirrel was very disappointed in him.
 'You are a terrible singer!' she yelled. 'I'm not giving you any of my wheat.'
 A tear rolled down Percy's cheek, and he vowed to give up singing for good.\\} 

True Analogy: \textit{Sam traveled all over the world buying beautiful things. When he ran out of money he paid a visit to his mother and gave her a gift he bought while in Tibet, hoping she would give him a loan in return. However, his mother was not at all pleased. 
 'You don't deserve any money of mine!' she exclaimed. 'This is a piece of junk!'\\}

False Analogy: \textit{Sam traveled all over the world buying beautiful things. When he ran out of money he paid a visit to his mother. However, she was not at all pleased with him.
 'While I have been hard at work you have been wasting your time,' she said. Sam gave her a gift he bought in Tibet, hoping she would give him a loan in return. But she was still not pleased. 'I will not give you any of my hard-earned money!' she exclaimed.\\} 

In this problem, the protagonist (Percy) expects something in return for his effort, but the result is a disappointment as the squirrel does not like what has been offered. Similarly, in both the target stories, the protagonist (Sam) presents a gift and expects something in return (i.e., money). However, the protagonist's mother rejects the offer and expresses disappointment. Critically, the reason of rejection is different in the two cases. One target story focuses on a negative judgment of the gift itself whereas the other emphasizes the mother's dissatisfaction with her son's behavior. Since both the stories have similar entities and first-order relations among them, the models have trouble with this difference. Particularly for the false analogy, it is possible the model loses track of why Percy's mother showed disappointment at first.

A similar analysis of problem 15, omitted here for reasons of space, suggests that the models might not be able to keep the track of the sequence of events. Finally, for problem 3, the false analogy appears to be a continuation of the plot of true analogy. Therefore, it is possible that the model's representation of the false analogy likely builds upon the true analogy representation, resulting in incorrect inferences. Lastly, for problem 13, the model seems to struggle to clearly determine how a character influences the subject's decision. Additionally, the structure and similarity of the analogies might have confused it, leading to a poor performance.  This post hoc analysis suggests that despite LLMs' competence at many tasks, they can fail for complex problems with conflicting analogies.
An aim for future work and model development is to address this limitation.


\section{Discussion}

\subsection{Summary of Findings}

Building on recent work by \citeA{Webb2023} and \citeA{Lewis2024EvaluatingTR}, we evaluated whether humans and LLMs show similar patterns of performance when understanding story analogies. Specifically, our experiments addressed four research questions. The first concerned the semantic representation of source and target stories. We encoded each story using BERT. The source story was more similar to the `true analogy' target than to the `false analogy' target 78\% of the time, which is slightly lower than the 84.7\% overall accuracy that humans achieve. However, a closer look revealed very similar representations of the `true analogy' and `false analogy' target stories in the model; see Figure~\ref{bert-similarity}. Moreover, and relevant for the third research question, there was no alignment between humans and BERT at the level of individual items.

The second research question was whether the alignment between humans and LLMs can be improved through a prompting strategy that uses self-generated hints. This was true for the overall accuracy measure: both GPT-4 and Llama 3.1 performed better for the enhanced prompt than the conventional prompt, with the larger versions of these models matching and even exceeding human performance; see Table \ref{gpt-llama-enhanced-prompt-performance}. Larger models can retain and utilize more intricate patterns within their training data, which can enable them to identify the nuanced relationships in story analogies. For example, a model with 70B parameters can effectively map causal relationships such as 'actions driven by revenge' and recognize subtle distinctions in sequences of events, while smaller models (e.g., with 8B parameters) might struggle with these deeper abstractions due to limited representational capacity.

However, there was no evidence that enhanced prompting enables LLMs to better align with humans at the level of individual items; see Figure \ref{correlation-matrix}. That said, this null finding may have been due to a ceiling effect: Performance of the larger models under enhanced prompting approached the maximum possible score (see Table \ref{gpt-llama-enhanced-prompt-performance}), reducing variability across items and therefore the ability to evaluate whether the models tracked human accuracies at this finer-grain level. In addition, a closer examination of the four problems that the models found much more difficult than humans did, suggested several features that may be difficult for LLMs to reason about; see Figure \ref{table_model_human_improved_prompt}.

The fourth research question concerned the effect of model size and architecture on the alignment of models to human analogical reasoning. Perhaps the most surprising result was that the alignment at the level of individual items was greater for the smaller models (i.e., GPT-4o-mini achieved a higher correlation with human accuracies than GPT-4o, and Llama-3.1-8B a higher correlation than Llama-3.1-70B); see Figure \ref{correlation-matrix}.


\subsection{Limitations and Future Directions}

This study takes a step beyond \citeA{Webb2023} and \citeA{Lewis2024EvaluatingTR} in evaluating whether current LLMs perform similarly to humans when reasoning about analogy stories. However, much work remains to be done, beyond continuing to evaluate the performance of future LLMs.

One limitation concerns the breadth of human data considered here -- or the lack thereof. We only considered data collected in the \citeA{Webb2023} replication of the \citeA{Gentner1993} study, which involved only 18 items. There is a rich cognitive science literature on analogical reasoning and problem solving that uses story materials, dating back to the seminal study of \citeA{gick1980analogical}, and before that \citeA{duncker1945problem}.
Future studies should include materials drawn from a broader range of studies. We note that obtaining the materials for older studies may be challenging, and the human performance data for individual items may be impossible. Replication studies may be necessary to assemble a comprehensive dataset on story analogy performance.

Another limitation of the current study may be the two prompting strategies that were employed. The enhanced prompt showed some advantages over the conventional prompt (i.e., overall accuracy was higher) but also some disadvantages (i.e., correlations to human performance at the item level were worse). Future studies should explore additional prompting techniques.

Finally, an important concern with testing large language models with published tasks is data contamination. Given the lack of transparency on the datasets on which generative models are being trained on, it is possible that these models are ``remembering'' the answers to these questions rather than reasoning about them. This may partly explain the increase in performance due to model size. However, further increases in performance due to the enhanced prompts show the importance of reasoning processes beyond rote memorization.


\section{Acknowledgements}
The first and second authors contributed equally to this paper.




\nocite{Webb2023}
\nocite{Gentner1993}
\nocite{Achiam2023GPT4TR}
\nocite{Touvron2023LLaMAOA}
\nocite{Devlin2019BERTPO}
\nocite{Lewis2024EvaluatingTR}
\nocite{gpt4}
\nocite{gpt4o-mini}
\nocite{llama3.1}
\nocite{Dubey2024TheL3}
\nocite{Brown2020LanguageMA}
\nocite{Sourati-etal-2024-arn}

\bibliographystyle{apacite}

\setlength{\bibleftmargin}{.125in}
\setlength{\bibindent}{-\bibleftmargin}

\bibliography{CogSci_Template}

\end{document}